\documentclass{article}

\usepackage[final]{neurips_2025}

\usepackage[utf8]{inputenc}
\usepackage[T1]{fontenc}
\usepackage{hyperref}
\usepackage{url}
\usepackage{booktabs}
\usepackage{amsfonts}
\usepackage{amsmath,amssymb,amsthm}
\usepackage{nicefrac}
\usepackage{microtype}
\usepackage{xcolor}
\usepackage{graphicx}
\usepackage{import}
\usepackage{enumitem}
\usepackage{tikz}
\usetikzlibrary{positioning,arrows.meta,shapes.geometric,calc,patterns,decorations.pathreplacing}

\newtheorem{proposition}{Proposition}

\newtheorem{definition}{Definition}

\DeclareMathOperator*{\argmin}{arg\,min}

\graphicspath{{Fig/}}

\title{\Large \bf Capability Thresholds and Manufacturing Topology:\\How Embodied Intelligence Triggers Phase Transitions\\in Economic Geography}

\author{%
Xinmin Fang \\
  LeTau Robotics\\
  University of Colorado Denver\\
  \texttt{xinmin.fang@ucdenver.edu} \\
  \And
  Lingfeng Tao \\
  LeTau Robotics\\
  Kennesaw State University\\
  \texttt{ltao2@kennesaw.edu} \\
  \AND
  Zhengxiong Li\thanks{Disclaimer: This working paper is intended solely for academic discussion and exploratory analysis. It does not constitute investment advice or serve as a basis for any financial decision-making. All data and examples referenced are derived from publicly available media sources and industry reports. \\This work is partially supported by the NVIDIA Academic Grant Program Award and US NSF Awards \#2426469, \#2426470, \#2436428, and \#2602731.} \\
  LeTau Robotics\\
  University of Colorado Denver\\
  \texttt{zhengxiong.li@ucdenver.edu} \\
}

\begin{document}

\maketitle

\begin{abstract}
The fundamental topology of manufacturing---where factories are built, how production is organized, at what scale it is viable---has not undergone a paradigm-level transformation since Henry Ford's moving assembly line in 1913. Every major innovation of the past century, from the Toyota Production System to Industry 4.0, has optimized \textit{within} the Fordist paradigm without altering its structural logic: centralized mega-factories, located near labor pools, producing at scale. We argue that embodied intelligence is poised to break this century-long stasis---not by making existing factories more efficient, but by triggering \textit{phase transitions} in manufacturing economic geography itself. When embodied AI capabilities cross critical thresholds in dexterity, task generalization, operational reliability, and tactile-vision fusion, the consequences extend far beyond cost reduction: they restructure where factories are built, how supply chains are organized, and what constitutes an economically viable production scale. We formalize this argument by defining a multi-dimensional \textit{Capability Space} $\mathcal{C} = (\delta, \gamma, \rho, \tau)$ and showing that the traditional manufacturing site-selection objective function undergoes topological reorganization when capability vectors cross specific critical surfaces. Through three transmission pathways---site-selection weight inversion, minimum economic batch collapse, and human-infrastructure decoupling---we demonstrate that embodied intelligence enables demand-proximal micro-manufacturing, eliminates the concept of ``manufacturing deserts,'' and potentially reverses decades of geographic concentration driven by labor arbitrage. We further introduce the concept of \textit{Machine Climate Advantage}: the observation that once human workers are removed from the equation, the optimal factory location is determined by machine-optimal environmental conditions (low humidity, high solar irradiance, thermal stability)---factors that are \textit{orthogonal} to traditional manufacturing siting logic, creating a geography of production with no historical precedent. This paper provides a first-principles framework at the intersection of robotics capability analysis and economic geography, establishing what we term \textit{Embodied Intelligence Economics}---the study of how physical AI capability thresholds reshape the spatial and structural logic of production.
\end{abstract}

\vspace{-0.5em}

\section{Introduction: Beyond the Efficiency Narrative}

The current discussion of embodied intelligence in manufacturing is dominated by an \textit{efficiency narrative}: robots replace human workers, production costs decline, throughput increases~\cite{duan2022survey,kroemer2021review}. This narrative, while not incorrect, is fundamentally incomplete. It treats embodied AI as a drop-in substitute within existing production systems---a faster, cheaper pair of hands on the same assembly line, in the same factory, in the same city chosen for the same reasons it was always chosen.

We argue that this framing misses the deeper structural transformation. Manufacturing site selection, supply chain topology, and production scale economics are not immutable constants of the industrial landscape---they are \textit{emergent properties} of the underlying capability constraints of the productive agents (human or robotic) that perform physical work~\cite{krugman1991geography,fujita1999spatial}. When these capabilities undergo qualitative improvement, the emergent economic structures must reorganize accordingly.

Consider a concrete example. A garment factory in Bangladesh exists in Bangladesh primarily because garment assembly requires dexterous manipulation of deformable materials---a task that, until recently, could only be performed by human hands~\cite{nayak2015automation}. The entire geographic logic of the global garment industry---$\$2.5$ trillion in annual output---is downstream of this single capability constraint~\cite{ILO2019garment}. If embodied AI systems achieve sufficient dexterity with deformable objects, the constraint dissolves, and with it the economic rationale for concentrating garment production in low-wage geographies.

This paper makes three contributions:

\begin{enumerate}[leftmargin=40pt,itemsep=2pt]
    \item We define a formal \textit{Capability Space} $\mathcal{C} = (\delta, \gamma, \rho, \tau)$ with four measurable dimensions---dexterity, generalization, reliability, and tactile-vision fusion---and argue that manufacturing economic geography is a function of position in this space.
    \item We identify three \textit{transmission pathways} through which capability improvements propagate into structural economic change: site-selection weight inversion, minimum economic batch collapse, and human-infrastructure decoupling.
    \item We introduce the concept of \textit{capability-triggered phase transitions}: critical surfaces in $\mathcal{C}$ beyond which the manufacturing topology undergoes discontinuous reorganization, analogous to thermodynamic phase transitions.
    \item We identify a novel siting logic---\textit{Machine Climate Advantage}---showing that beyond full autonomy, optimal factory locations are determined by machine-optimal environmental conditions (low humidity, high irradiance, thermal stability) that are orthogonal to traditional manufacturing siting factors, opening a geography of production with no historical precedent.
\end{enumerate}

To our knowledge, this is the first work to formally bridge embodied AI capability analysis with manufacturing economic geography through a phase-transition framework. We term this intersection \textit{Embodied Intelligence Economics}. We begin by arguing (Section~2) that this potential paradigm shift must be understood against a striking historical backdrop: the fundamental topology of manufacturing has not changed since Henry Ford's assembly line in 1913---over a century of frozen paradigm.

\section{A Century of Frozen Topology: From Highland Park to Industry 4.0}

Before formalizing our framework, it is worth pausing to appreciate the historical anomaly we are confronting. The fundamental topology of manufacturing---the spatial logic of where factories are built, how production is organized, and what determines optimal scale---has not undergone a paradigm-level transformation in over a century. Every major development since Henry Ford's Highland Park plant in 1913 has been an \textit{optimization within} the Fordist paradigm, not a departure from it.

\subsection{The Fordist Paradigm: The Last True Phase Transition}

When Ford introduced the moving assembly line for the Model T in 1913, he did not merely improve an existing process. He restructured the \textit{topology} of production itself~\cite{hounshell1984american,ford1926today}. Prior to Ford, manufacturing was organized around craft workshops---skilled artisans performed complete assemblies at individual stations, and production scaled by adding more artisans. Ford's insight was to decompose complex assembly into elementary, sequential operations performed by low-skilled workers at fixed stations along a moving line.

The consequences were paradigmatic, not incremental:

\begin{itemize}[leftmargin=40pt,itemsep=2pt]
    \item \textbf{Scale economics:} The assembly line created enormous fixed costs (tooling, line design, facility layout) that could only be amortized over massive production runs, establishing the economic logic of centralized mega-factories.
    \item \textbf{Labor deskilling:} By decomposing tasks to their elementary constituents, Ford decoupled production from the geographic distribution of skilled labor, enabling factories wherever unskilled labor pools existed~\cite{braverman1974labor}.
    \item \textbf{Geographic concentration:} The combination of scale economics, transportation networks, and access to large labor pools produced the industrial agglomerations that still define the manufacturing map today---Detroit, the Ruhr Valley, the Pearl River Delta~\cite{krugman1991geography}.
\end{itemize}

This was a genuine phase transition in manufacturing topology: a discontinuous reorganization of the spatial and structural logic of production.

\subsection{A Century of Optimization Within the Paradigm}

Remarkably, every major manufacturing innovation since 1913 has operated \textit{within} the topological framework Ford established, refining its parameters without altering its fundamental structure. Table~\ref{tab:century} summarizes this trajectory.

\begin{table}[hbt!]
\centering
\caption{A century of manufacturing innovations: paradigm optimization, not paradigm change}
\label{tab:century}
\small
\begin{tabular}{lcp{5.5cm}c}
\toprule
\textbf{Innovation} & \textbf{Era} & \textbf{What Changed} & \textbf{Topology Changed?} \\
\midrule
Moving assembly line & 1913 & Sequential task decomposition at scale & \textbf{Yes} \\
\hline
Statistical quality control & 1940s & Defect reduction via measurement & No \\
\hline
Toyota Production System & 1960s & Waste elimination, just-in-time flow & No \\
\hline
CNC machining & 1970s & Programmable precision cutting & No \\
\hline
Industrial robotics & 1980s & Automated spot welding, painting & No \\
\hline
Lean manufacturing & 1990s & Value-stream mapping, pull systems & No \\
\hline
ERP/MES systems & 2000s & Digital coordination of production & No \\
\hline
Industry 4.0 / IoT & 2010s & Sensor networks, digital twins & No \\
\hline
Collaborative robots & 2020s & Flexible human-robot cowork & No \\
\bottomrule
\end{tabular}
\end{table}

Consider the specifics. The Toyota Production System (TPS), widely regarded as the most significant manufacturing innovation since Ford~\cite{ohno1988toyota,womack1990machine}, transformed \textit{how} work flows through a factory (pull vs.\ push, kanban, heijunka), but the factory itself remained a centralized facility, located near labor pools and transportation hubs, producing at scale. Lean manufacturing compressed inventory and reduced waste but did not alter the fundamental equation: large facilities, human workers, labor-proximal location.

Even Industry 4.0---the digitization of manufacturing through IoT sensors, digital twins, and cyber-physical systems~\cite{kagermann2013industry40}---represents a \textit{nervous system upgrade} to the Fordist body, not a new body. The factory is smarter, more instrumented, more responsive---but it is still a large, centralized, human-staffed facility whose location is determined by the same factors that determined Ford's choice of Highland Park.

Industrial robots, first deployed at General Motors in 1961 (Unimate) and now numbering over 4 million globally~\cite{IFR2024report}, automate \textit{specific stations} within the assembly line but do not eliminate the human workforce or the infrastructure it requires. A modern automotive plant with 1,000 robots still employs 3,000--5,000 human workers~\cite{smil2017energy}. The topology is unchanged: the robot is a faster, more consistent worker on Ford's line, not an alternative to the line itself.

\subsection{Why Now? The Qualitative Discontinuity of Embodied Intelligence}

What makes embodied intelligence different from the optimizations of the past century? We argue it is the convergence of three capabilities that, together, attack the foundational assumptions of the Fordist paradigm simultaneously:

\begin{enumerate}[leftmargin=40pt,itemsep=2pt]
    \item \textbf{Foundation models eliminate task specificity.} Traditional industrial robots are programmed for a single task; changing the task requires reprogramming, retooling, and recalibration. Foundation models for robotic manipulation~\cite{brohan2023rt2,black2024pi0} enable few-shot or zero-shot adaptation to novel tasks, breaking the assumption that automation requires massive fixed investment per task---the very assumption that drives scale economics and centralization.
    \item \textbf{Dexterous manipulation closes the last-mile gap.} For a century, the ``last 30\%'' of manufacturing tasks---those requiring dexterous handling of deformable, fragile, or geometrically complex objects---have remained stubbornly human. Recent advances in dexterous hand design, tactile sensing, and sim-to-real transfer~\cite{zhang2025dexterous,chen2023visual} are closing this gap. When it closes, the requirement for human workers---and thus for human infrastructure---dissolves.
    \item \textbf{Reliability at the 99.9\%+ level enables autonomy.} Previous automation achieved high reliability \textit{for specific, repetitive tasks} but required human supervision for exception handling. The combination of foundation-model generalization and robust perception allows embodied systems to handle exceptions autonomously, eliminating the human-in-the-loop requirement and, with it, the spatial coupling between factory and labor pool.
\end{enumerate}

It is the \textit{convergence} of these three capabilities---not any one alone---that constitutes a paradigmatic challenge to the Fordist topology. Task-specific automation (even if highly dexterous) merely optimizes existing lines. Generalizable AI (without physical embodiment) cannot perform manufacturing tasks. Only the combination of generalization, dexterity, and autonomous reliability creates the conditions for a new phase transition in manufacturing topology.

The rest of this paper formalizes this argument. We define a multi-dimensional Capability Space that captures these convergent capabilities, identify the critical thresholds at which they trigger structural change, and characterize the new manufacturing topologies that emerge beyond these thresholds. If we are correct, the coming decade will witness the first genuine paradigm shift in the spatial organization of manufacturing since Henry Ford's Highland Park plant opened its doors in 1913.

\section{The Multi-Dimensional Capability Space}

Existing analyses of automation's economic impact typically reduce robot capability to a single scalar---``automation level'' or ``robot density''~\cite{acemoglu2020robots,graetz2018robots}. This is analytically convenient but obscures the fact that different \textit{dimensions} of capability have qualitatively different economic consequences. A robot that is 99\% reliable but cannot handle deformable objects has fundamentally different economic implications than one that handles deformable objects but fails 10\% of the time.

We propose a four-dimensional \textbf{Capability Space} $\mathcal{C}$:

\begin{definition}[Capability Space]
The Embodied Intelligence Capability Space is the product space $\mathcal{C} = \mathcal{D} \times \mathcal{G} \times \mathcal{R} \times \mathcal{T}$, where:
\begin{itemize}[leftmargin=40pt,itemsep=2pt]
    \item $\delta \in \mathcal{D} = [0,1]$: \textbf{Dexterity Index}---the fraction of manipulation tasks in a target industry achievable by the embodied system, weighted by economic value.
    \item $\gamma \in \mathcal{G} = [0,1]$: \textbf{Generalization Index}---the probability that the system successfully performs a novel task variant with $\leq k$ demonstrations (few-shot adaptation).
    \item $\rho \in \mathcal{R} = [0,1]$: \textbf{Reliability Index}---the probability of continuous successful operation over a standard shift duration without human intervention.
    \item $\tau \in \mathcal{T} = [0,1]$: \textbf{Tactile-Vision Fusion Index}---the fraction of quality-critical inspection and manipulation tasks achievable through integrated tactile and visual sensing.
\end{itemize}
\end{definition}

A capability state is a vector $\mathbf{c} = (\delta, \gamma, \rho, \tau) \in \mathcal{C}$. The current state of embodied AI for most manufacturing domains occupies the lower-left region of this space. The central thesis of this paper is that there exist \textit{critical surfaces} $\Sigma_i \subset \mathcal{C}$ such that crossing $\Sigma_i$ triggers a qualitative reorganization of manufacturing economic structure.

\subsection{Dimension 1: Dexterity ($\delta$)}

Dexterity is the most intuitive capability dimension and the one most extensively studied in the robotics literature~\cite{kroemer2021review,zhang2025dexterous}. For our purposes, what matters is not the absolute manipulation capability but the \textit{economic coverage}---what fraction of value-creating manipulation tasks in a given industry can the system perform?

Consider electronics assembly. Current industrial robots handle pick-and-place of rigid components with high success rates ($\delta_{\text{rigid}} \approx 0.95$)~\cite{IFR2024report}. However, tasks involving flexible cables, connectors with sub-millimeter tolerances, or multi-step assembly sequences with force-sensitive insertion remain challenging ($\delta_{\text{flex}} \approx 0.3$--$0.5$)~\cite{zhang2025dexterous}. These latter tasks account for a disproportionate share of manual labor in electronics manufacturing.

The economic significance of $\delta$ is nonlinear. A factory that can automate 70\% of tasks still requires a full human workforce infrastructure (break rooms, parking, HVAC for comfort, shift scheduling, safety protocols) to support the remaining 30\%. Only when $\delta$ approaches a critical value $\delta^*$---which we estimate to be industry-specific and typically in the range $[0.90, 0.98]$---does the ``last mile'' of human labor become eliminable, and with it the requirement for human-centric infrastructure.

\subsection{Dimension 2: Generalization ($\gamma$)}

Generalization captures the ability of embodied AI systems to adapt to new tasks without extensive reprogramming or retraining~\cite{brohan2023rt2,black2024pi0}. This dimension has undergone the most dramatic recent improvement due to foundation models for robotics.

The economic consequence of $\gamma$ is felt primarily through \textit{production flexibility}. Traditional industrial automation achieves high throughput for fixed tasks but incurs substantial switching costs: retooling time, reprogramming, and recalibration can take days to weeks~\cite{koren2010globalmanufacturing}. This creates strong economies of scale---large production runs amortize switching costs, which in turn demands centralized, high-volume facilities.

If $\gamma$ is high---meaning an embodied system can adapt to a new product variant in minutes rather than days---then the switching cost approaches zero. This has a cascading effect on optimal production scale:

\begin{equation}
\text{Minimum Economic Batch Size} = \frac{C_{\text{fixed}} + C_{\text{switch}}(\gamma)}{p - c_{\text{marginal}}}
\label{eq:batch}
\end{equation}

where $C_{\text{switch}}(\gamma)$ is a monotonically decreasing function of $\gamma$, approaching zero as $\gamma \to 1$. When the minimum economic batch size drops below a critical threshold, centralized mass production loses its structural advantage over distributed, demand-proximal manufacturing.

\subsection{Dimension 3: Reliability ($\rho$)}

Reliability determines whether embodied systems can operate \textit{autonomously} or require human supervision. This distinction is economically critical because supervision requirements maintain the spatial coupling between manufacturing and human labor pools~\cite{parasuraman2000model}.

The relationship between $\rho$ and required human oversight is sharply nonlinear. At $\rho = 0.99$ (one failure per 100 operations), a production line of 50 stations experiences, on average, a failure every 2 operations---requiring near-continuous human monitoring. At $\rho = 0.999$, the same line fails once every 20 cycles, allowing intermittent supervision. At $\rho = 0.9999$, failures become rare enough to be handled by remote exception management rather than on-site personnel.

For a production line of $n$ stations with per-station reliability $\rho$, the line-level yield (probability of a full cycle without failure) is:
\begin{equation}
Y_{\text{line}} = \rho^n
\label{eq:yield}
\end{equation}

For $n = 50$: $Y_{\text{line}}(0.99) = 0.605$; $Y_{\text{line}}(0.999) = 0.951$; $Y_{\text{line}}(0.9999) = 0.995$. The transition from ``requires on-site human workforce'' to ``remotely manageable'' occurs in a narrow band of $\rho$, creating a natural phase transition in staffing requirements and, consequently, in site-selection constraints.

\subsection{Dimension 4: Tactile-Vision Fusion ($\tau$)}

The final dimension captures the integration of tactile sensing with visual perception~\cite{luo2017robotic,li2020review}. Many manufacturing tasks that are trivial for humans---assessing material quality by touch, detecting assembly errors through force feedback, handling objects with variable compliance---remain challenging for vision-only robotic systems.

Industries with high $\tau$ requirements include food processing (ripeness assessment, gentle handling), textile manufacturing (fabric manipulation, seam quality), pharmaceutical packaging (blister pack verification, vial integrity), and precision assembly (connector insertion with force limits)~\cite{dahiya2010tactile}. These industries collectively employ hundreds of millions of workers globally and are disproportionately concentrated in low-wage economies precisely because the required tactile intelligence has resisted automation.

The economic significance of $\tau$ is thus heavily concentrated in labor-intensive industries where the geographic arbitrage motive is strongest. Advances in $\tau$ disproportionately affect the industries whose location logic is most dependent on cheap labor.

\section{Three Transmission Pathways}

How do improvements in $\mathbf{c} = (\delta, \gamma, \rho, \tau)$ propagate into changes in manufacturing economic geography? We identify three distinct transmission pathways.

\subsection{Pathway 1: Site-Selection Weight Inversion}

Manufacturing site selection is classically modeled as a multi-criteria optimization problem~\cite{yang2007facility,weber1929theory}:

\begin{equation}
\mathbf{x}^* = \argmin_{\mathbf{x} \in \mathcal{X}} \sum_{i} w_i(\mathbf{c}) \cdot f_i(\mathbf{x})
\label{eq:site}
\end{equation}

where $\mathbf{x}$ is a candidate location, $f_i(\mathbf{x})$ are cost functions for factors such as labor ($f_L$), logistics ($f_T$), land ($f_G$), energy ($f_E$), market proximity ($f_M$), and regulatory environment ($f_R$), and $w_i(\mathbf{c})$ are weights that depend on the capability state $\mathbf{c}$.

The critical insight is that the weights $w_i$ are not constants---they are functions of embodied AI capability. Specifically:

\begin{equation}
w_L(\mathbf{c}) = w_L^{(0)} \cdot \left(1 - \delta \cdot \rho \cdot h(\gamma, \tau)\right)
\label{eq:labor_weight}
\end{equation}

where $w_L^{(0)}$ is the baseline labor weight and $h(\gamma, \tau) \in [0,1]$ captures the joint contribution of generalization and tactile-vision capability. As $\mathbf{c}$ improves, $w_L$ decreases, and the relative importance of other factors---particularly $w_M$ (market proximity) and $w_E$ (energy cost)---increases.

\begin{proposition}[Weight Inversion]
There exists a critical capability surface $\Sigma_W \subset \mathcal{C}$ such that for $\mathbf{c}$ below $\Sigma_W$, $w_L(\mathbf{c}) > w_M(\mathbf{c})$ (labor cost dominates market proximity), and for $\mathbf{c}$ above $\Sigma_W$, $w_M(\mathbf{c}) > w_L(\mathbf{c})$ (market proximity dominates labor cost). Crossing $\Sigma_W$ triggers a discontinuous shift in the Pareto-optimal location set $\mathcal{X}^*$.
\end{proposition}

The consequence is a \textit{topological reorganization} of the optimal manufacturing map. Locations that were never competitive under labor-dominated weighting---city centers in high-wage economies, remote areas with cheap energy but no labor pool, or regions close to end consumers but with high labor costs---suddenly enter the Pareto-optimal set. Conversely, locations whose primary advantage was cheap labor may exit the optimal set.

This is not a gradual shift. Because the Pareto frontier is defined by the intersection of cost iso-surfaces, a change in weight ordering can cause a discontinuous jump in the optimal location set---mathematically analogous to a first-order phase transition in thermodynamics~\cite{goldenfeld1992lectures}.

\subsection{Pathway 2: Minimum Economic Batch Collapse}

The second pathway operates through production scale economics. The minimum economic batch size (MEBS) is the smallest production run for which revenue exceeds total cost including amortized fixed costs:

\begin{equation}
\text{MEBS}(\mathbf{c}) = \frac{C_{\text{facility}} + C_{\text{equip}} + C_{\text{switch}}(\gamma) + C_{\text{labor}}(\delta, \rho)}{p - c_{\text{var}}}
\label{eq:mebs}
\end{equation}

Foundation models for robotic manipulation~\cite{brohan2023rt2,black2024pi0} directly attack $C_{\text{switch}}(\gamma)$ by enabling rapid task adaptation without physical retooling. Dexterity and reliability improvements reduce $C_{\text{labor}}(\delta, \rho)$ by eliminating the need for human workers and supervisors.

As both terms shrink, the MEBS decreases. When MEBS drops below a critical value $N^*$, a qualitative shift occurs in optimal production topology:

\begin{proposition}[Topology Transition]
Define $N^*$ as the demand volume of a single metropolitan market for a given product category. When $\text{MEBS}(\mathbf{c}) > N^*$, the cost-minimizing production topology is centralized (one or few large facilities serving global markets). When $\text{MEBS}(\mathbf{c}) < N^*$, the cost-minimizing topology is distributed (many small facilities each serving local markets). The transition at $\text{MEBS} = N^*$ is sharp because it coincides with the elimination of logistics costs for inter-regional distribution.
\end{proposition}

This transition has profound implications for supply chains. In the centralized regime, supply chains are long, inventories are large, and lead times are measured in weeks. In the distributed regime, supply chains collapse to local sourcing, inventories approach zero (make-to-order becomes feasible), and lead times are measured in hours. The entire $\$10$ trillion global logistics industry~\cite{statista2024logistics} is structured around the centralized regime; the distributed regime would require---and enable---a fundamentally different logistics architecture.

\subsection{Pathway 3: Human-Infrastructure Decoupling}

The third pathway is perhaps the most radical. Current manufacturing locations are constrained not only by labor \textit{cost} but by labor \textit{availability}---which requires an entire ecosystem of human infrastructure: housing, transportation, healthcare, education, sanitation, food supply, and social services~\cite{glaeser2011triumph}. A factory cannot be built where workers cannot live.

This constraint creates what we term \textit{manufacturing deserts}---regions that are geographically or strategically advantageous for production but lack the human-infrastructure base to support a manufacturing workforce. Examples include:

\begin{itemize}[leftmargin=40pt,itemsep=2pt]
    \item Arctic and sub-Arctic regions with abundant renewable energy (geothermal, wind) but sparse populations
    \item Arid regions proximal to mineral resources but lacking water and habitation infrastructure
    \item Offshore platforms and artificial islands near major shipping lanes
    \item Space-based or lunar manufacturing for in-situ resource utilization~\cite{yoshida2003achievements}
\end{itemize}

When the capability vector $\mathbf{c}$ crosses the critical surface $\Sigma_H$ defined by:
\begin{equation}
\Sigma_H: \quad \delta \cdot \rho > \theta_H \quad \text{and} \quad \gamma > \theta_G
\label{eq:decoupling}
\end{equation}

where $\theta_H$ and $\theta_G$ are industry-specific thresholds, the factory becomes fully autonomous---requiring only energy, raw materials, computation, and periodic remote maintenance. The ``manufacturing desert'' concept dissolves, and the set of feasible manufacturing locations expands discontinuously from the set of \textit{habitable} regions to the set of \textit{energy-accessible} regions.

\begin{proposition}[Feasibility Set Expansion]
Let $\mathcal{L}_{\text{hab}}$ denote the set of locations with sufficient human infrastructure and $\mathcal{L}_{\text{energy}}$ denote the set of locations with sufficient energy access. Currently, feasible manufacturing locations satisfy $\mathcal{L}_{\text{feasible}} \subseteq \mathcal{L}_{\text{hab}} \cap \mathcal{L}_{\text{energy}}$. Beyond $\Sigma_H$, $\mathcal{L}_{\text{feasible}} = \mathcal{L}_{\text{energy}}$. Since $\mathcal{L}_{\text{energy}} \supsetneq \mathcal{L}_{\text{hab}}$, this represents a strict expansion of the feasible manufacturing set.
\end{proposition}

\subsubsection{Machine Climate Advantage: A New Siting Logic}

The decoupling pathway has a further consequence that, to our knowledge, has not been recognized in the existing literature. Once human presence is no longer required, the optimal factory location is determined not by traditional manufacturing siting factors (labor cost, labor availability, port access) but by \textit{machine-optimal environmental conditions}---what we term the \textbf{Machine Climate Advantage (MCA)}.

The key observation is that robotic systems and human workers have \textit{fundamentally different} environmental preferences---and crucially, the factors that matter most to machines have been entirely absent from a century of manufacturing site-selection theory. Humans require narrow temperature bands (20--25$^\circ$C), moderate humidity (40--60\%), natural light for psychological wellbeing, clean air, and proximity to healthcare, education, and social infrastructure~\cite{glaeser2011triumph}. These factors have dominated industrial planning. Robotic systems, by contrast, are optimized by an entirely different set of environmental parameters:

\begin{itemize}[leftmargin=40pt,itemsep=2pt]
    \item \textbf{Low humidity} ($<$30\% RH): reduces corrosion of metallic joints and actuators, prevents condensation on optical sensors and cameras, minimizes oxidation of electronic components, and eliminates mold risk on stored materials~\cite{lee2020corrosion}.
    \item \textbf{Low and stable dust levels}: extends the operational lifetime of precision bearings, optical systems (LiDAR, stereo cameras), and tactile sensor arrays.
    \item \textbf{Temperature stability over absolute range}: robots tolerate wide temperature ranges ($-20^\circ$C to $+50^\circ$C for most industrial systems) but are sensitive to rapid thermal cycling that induces mechanical stress and sensor drift. Low diurnal humidity variation matters more than absolute temperature.
    \item \textbf{High solar irradiance}: enables cost-effective solar energy generation for autonomous operation; provides consistent, high-quality natural illumination that improves vision system performance and reduces reliance on artificial lighting.
    \item \textbf{Low precipitation frequency}: reduces operational interruptions for facilities with outdoor or semi-outdoor logistics, loading, and material handling zones.
    \item \textbf{Moderate altitude}: lower air density at altitude reduces aerodynamic drag on high-speed manipulators and improves convective cooling efficiency for electronics, though the effect is second-order.
\end{itemize}

We formalize this through an \textbf{Environmental Adaptation Function} $\phi(\mathbf{x})$ that modulates the effective reliability of an embodied system as a function of its operating location $\mathbf{x}$:

\begin{equation}
\rho_{\text{eff}}(\mathbf{x}) = \rho_{\text{base}} \cdot \phi(\mathbf{x}), \qquad \phi(\mathbf{x}) = \prod_{j} \phi_j(e_j(\mathbf{x}))
\label{eq:phi}
\end{equation}

where $e_j(\mathbf{x})$ are environmental parameters at location $\mathbf{x}$ (humidity, dust, thermal stability, solar irradiance, precipitation frequency) and $\phi_j: \mathbb{R} \to (0, 1]$ are monotonic response functions for each parameter, normalized so that $\phi_j = 1$ at the machine-optimal value. The product structure reflects the fact that reliability degradation from multiple environmental stressors compounds multiplicatively~\cite{pecht2009product}.

The critical implication is that $\phi(\mathbf{x})$ and traditional manufacturing siting attractiveness $S(\mathbf{x})$ are \textit{not positively correlated}---and in many cases are \textit{orthogonal}. The environmental factors that maximize $\phi$ (aridity, high irradiance, low precipitation, thermal stability) have simply been \textit{irrelevant} to manufacturing site selection throughout industrial history, because they do not benefit human workers. A location can be perfectly livable---even highly desirable---for human residents while simultaneously possessing excellent machine-climate characteristics, yet still have been overlooked for manufacturing because it lacked the traditional prerequisites: port access, large blue-collar labor pools, heavy-industry heritage, or low wage costs. This creates a new category of location advantage that we term \textbf{Machine Climate Advantage}:

\begin{definition}[Machine Climate Advantage]
A location $\mathbf{x}$ possesses Machine Climate Advantage if $\phi(\mathbf{x}) > \phi(\mathbf{x}')$ for typical current manufacturing locations $\mathbf{x}'$, even though $S(\mathbf{x}) < S(\mathbf{x}')$ under traditional (labor-dependent) siting criteria. That is, $\mathbf{x}$ is a superior location for autonomous manufacturing precisely because it possesses environmental attributes---aridity, high irradiance, thermal stability---that were invisible to the traditional site-selection calculus.
\end{definition}

This redefines the site-selection objective function (Equation~\ref{eq:site}) beyond $\Sigma_H$. The labor weight $w_L$ drops to zero, and a new term emerges:

\begin{equation}
\mathbf{x}^* = \argmin_{\mathbf{x}} \Big[ w_E \cdot f_E(\mathbf{x}) + w_M \cdot f_M(\mathbf{x}) + w_T \cdot f_T(\mathbf{x}) + w_\phi \cdot \big(1 - \phi(\mathbf{x})\big) \Big]
\label{eq:site_mca}
\end{equation}

where $w_\phi \cdot (1 - \phi(\mathbf{x}))$ penalizes locations with suboptimal machine environments. The weight $w_\phi$ increases with the scale of autonomous operations, as the cumulative MTBF (mean time between failures) gain from operating in a machine-optimal climate compounds across thousands of robotic units running continuously.

This represents a \textit{third logic} of manufacturing site selection, fundamentally distinct from both the labor-proximal logic (Phase~I) and the market-proximal logic (Phase~II):

\begin{itemize}[leftmargin=40pt,itemsep=2pt]
    \item \textbf{Phase I:} Factories go where workers are cheap.
    \item \textbf{Phase II:} Factories go where customers are close.
    \item \textbf{Phase III:} Factories go where machines run best.
\end{itemize}

\section{Phase Transition Framework}

We now formalize the notion of capability-triggered phase transitions by defining an order parameter for manufacturing topology.

\subsection{The Manufacturing Concentration Index}

Let $\mathcal{M}(\mathbf{c})$ denote the spatial distribution of manufacturing activity as a function of the capability state $\mathbf{c}$. We define the \textbf{Manufacturing Concentration Index} (MCI) as:

\begin{equation}
\text{MCI}(\mathbf{c}) = \sum_{r \in \mathcal{R}} \left(\frac{m_r(\mathbf{c})}{M(\mathbf{c})}\right)^2
\label{eq:mci}
\end{equation}

where $m_r(\mathbf{c})$ is manufacturing output in region $r$ and $M(\mathbf{c}) = \sum_r m_r(\mathbf{c})$ is total global output. MCI $= 1$ corresponds to complete concentration (all manufacturing in one region); MCI $= 1/|\mathcal{R}|$ corresponds to uniform distribution.

\subsection{The Phase Diagram}

We conjecture that MCI exhibits discontinuous behavior as a function of $\mathbf{c}$, analogous to a first-order phase transition:

\begin{equation}
\text{MCI}(\mathbf{c}) = 
\begin{cases}
\text{MCI}_{\text{concentrated}} & \text{if } \mathbf{c} \in \mathcal{C}_{\text{below}} \\
\text{MCI}_{\text{distributed}} & \text{if } \mathbf{c} \in \mathcal{C}_{\text{above}}
\end{cases}
\label{eq:phase}
\end{equation}

where the transition occurs at the critical surface $\Sigma = \Sigma_W \cap \Sigma_N \cap \Sigma_H$ (the intersection of the three pathway thresholds). The ``order parameter'' MCI undergoes a discontinuous drop at $\Sigma$, driven by the simultaneous activation of all three transmission pathways.

\begin{figure}[hbt!]
\centering
\begin{tikzpicture}[scale=1.0]

\fill[blue!6] (0,0) rectangle (11,8);
\fill[orange!8] (4.2,8) .. controls (5.2,5.8) and (6,4.2) .. (11,2.6) -- (11,8) -- cycle;
\fill[green!10] (8.2,8) .. controls (8.8,6.8) and (9.5,5.8) .. (11,5.2) -- (11,8) -- cycle;

\foreach \x in {2,4,6,8,10} {
    \draw[gray!15, thin] (\x,0) -- (\x,8);
}
\foreach \y in {1.6,3.2,4.8,6.4,8} {
    \draw[gray!15, thin] (0,\y) -- (11,\y);
}

\draw[-{Stealth[length=3mm]}, thick] (0,0) -- (11.5,0);
\draw[-{Stealth[length=3mm]}, thick] (0,0) -- (0,8.8);
\node[font=\footnotesize, anchor=north] at (5.5,-0.6) {Dexterity $\times$ Reliability \;($\delta \cdot \rho$)};
\node[font=\footnotesize, anchor=south, rotate=90] at (-0.9,4) {Generalization \;($\gamma$)};

\foreach \x/\lab in {2/0.2, 4/0.4, 6/0.6, 8/0.8, 10/1.0} {
    \draw (\x,0) -- (\x,-0.15) node[below, font=\scriptsize] {\lab};
}
\foreach \y/\lab in {1.6/0.2, 3.2/0.4, 4.8/0.6, 6.4/0.8, 8/1.0} {
    \draw (0,\y) -- (-0.15,\y) node[left, font=\scriptsize] {\lab};
}

\draw[blue!70!black, very thick, dashed] (4.2,8) .. controls (5.2,5.8) and (6,4.2) .. (11,2.6);
\draw[red!70!black, very thick, densely dotted] (6.2,8) .. controls (7,6) and (7.8,5) .. (11,3.8);
\draw[green!50!black, very thick, dashdotted] (8.2,8) .. controls (8.8,6.8) and (9.5,5.8) .. (11,5.2);

\node[font=\scriptsize\bfseries, blue!70!black, anchor=west] at (11.15,2.6) {$\Sigma_W$};
\node[font=\scriptsize\bfseries, red!70!black, anchor=west] at (11.15,3.8) {$\Sigma_N$};
\node[font=\scriptsize\bfseries, green!50!black, anchor=west] at (11.15,5.2) {$\Sigma_H$};

\node[draw=gray!60, fill=white, rounded corners=3pt, inner sep=5pt,
      font=\scriptsize, anchor=north west, align=left] at (0.3,8.5) {
    \raisebox{1pt}{\tikz\draw[blue!70!black, very thick, dashed] (0,0)--(0.6,0);}\; $\Sigma_W$: Weight Inversion \\[2pt]
    \raisebox{1pt}{\tikz\draw[red!70!black, very thick, densely dotted] (0,0)--(0.6,0);}\; $\Sigma_N$: Batch Collapse \\[2pt]
    \raisebox{1pt}{\tikz\draw[green!50!black, very thick, dashdotted] (0,0)--(0.6,0);}\; $\Sigma_H$: Decoupling
};

\node[font=\small, align=center, text=blue!60!black] at (2.2,2.8) {
    \textbf{Phase I}\\[2pt]{\footnotesize Centralized}\\[-1pt]{\footnotesize Labor-Proximal}\\[-1pt]{\footnotesize Manufacturing}};
\node[font=\small, align=center, text=red!60!black] at (8.8,1.5) {
    \textbf{Phase II}\\[2pt]{\footnotesize Market-Proximal}\\[-1pt]{\footnotesize Distributed}\\[-1pt]{\footnotesize Manufacturing}};
\node[font=\small, align=center, text=green!45!black] at (9.8,7.2) {
    \textbf{Phase III}\\[2pt]{\footnotesize Ubiquitous}\\[-1pt]{\footnotesize Autonomous}\\[-1pt]{\footnotesize Manufacturing}};

\filldraw[black] (2.0,1.4) circle (3pt);
\node[font=\scriptsize, anchor=north, yshift=-2pt] at (2.0,1.4) {Current State};

\draw[-{Stealth[length=2.5mm]}, thick, gray!70, dashed] (2.0,1.4) -- (8.2,5.8);
\node[font=\scriptsize, gray!70, rotate=35, anchor=north] at (3.8,2.8) {Technology Trajectory};

\filldraw[blue!70!black] (5.6,3.5) circle (2.5pt);
\node[font=\tiny, blue!70!black, anchor=north west, xshift=3pt, yshift=-1pt] at (5.6,3.5) {Electronics};
\filldraw[red!70!black] (6.8,4.3) circle (2.5pt);
\node[font=\tiny, red!70!black, anchor=north west, xshift=2pt, yshift=-1pt] at (6.8,4.3) {Food};
\filldraw[green!50!black] (7.6,5.4) circle (2.5pt);
\node[font=\tiny, green!50!black, anchor=south east, xshift=-2pt, yshift=1pt] at (7.6,5.4) {Aerospace};

\end{tikzpicture}
\caption{Conceptual phase diagram of manufacturing topology in the $(\delta \cdot \rho, \gamma)$ projection of Capability Space $\mathcal{C}$. Three critical surfaces separate distinct manufacturing regimes. \textbf{Phase~I} (current): labor cost dominates site selection, production is centralized, and facilities require human infrastructure. \textbf{Phase~II}: after crossing $\Sigma_W$ (weight inversion), market proximity dominates and distributed manufacturing becomes optimal. \textbf{Phase~III}: beyond $\Sigma_H$ (decoupling), manufacturing is viable wherever energy exists, independent of human habitation. The dashed arrow indicates the projected technology trajectory based on current foundation model scaling trends.}
\label{fig:phase}
\end{figure}
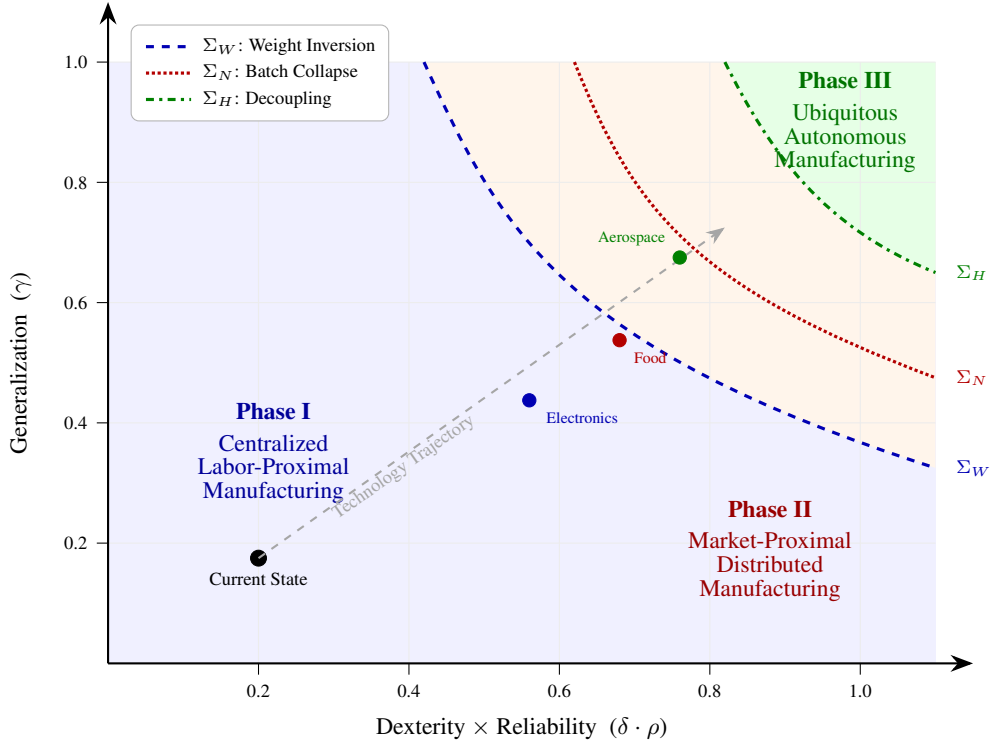

Figure~\ref{fig:phase} illustrates the conceptual phase diagram in a two-dimensional projection of $\mathcal{C}$.

\section{Industry Case Studies: Three Paths to Phase Transition}

The critical surfaces $\Sigma_i$ are not universal---they vary substantially by industry, depending on the specific manipulation challenges, current automation levels, supply chain structures, and geographic concentration drivers. We present three detailed case studies that illustrate how the Capability Space framework maps onto concrete industrial transformations. Each case is chosen to highlight a different primary transmission pathway: electronics assembly ($\Sigma_W$: weight inversion), aerospace manufacturing ($\Sigma_N$: batch collapse), and food processing ($\Sigma_H$: human-infrastructure decoupling).

\subsection{Case 1: Consumer Electronics Assembly --- The Weight Inversion Trigger}

\subsubsection{Current Landscape}

The consumer electronics industry---smartphones, laptops, wearables---represents a \$1.1 trillion annual market, with over 80\% of final assembly concentrated in China and Southeast Asia~\cite{IFR2024report}. The assembly of a single smartphone involves approximately 70--80 discrete manipulation steps~\cite{flexport2023electronics}, of which roughly 50--55 are currently performed by human workers. The automated steps (SMT placement, solder reflow, optical inspection) involve rigid components on flat substrates; the manual steps involve:

\begin{itemize}[leftmargin=40pt,itemsep=2pt]
    \item \textbf{Flexible cable routing and connection:} FPC (flexible printed circuit) cables must be threaded through tight channels and snapped into ZIF (zero insertion force) connectors with sub-millimeter alignment and controlled insertion force~\cite{foxconn2023challenges}.
    \item \textbf{Battery insertion and adhesive management:} Pouch-cell batteries are deformable, force-sensitive, and require precise adhesive placement for thermal management.
    \item \textbf{Gasket and seal placement:} Waterproofing requires precise placement of deformable rubber gaskets in complex 3D geometries.
    \item \textbf{Final assembly and cosmetic inspection:} Display bonding, back cover attachment, and cosmetic quality verification require simultaneous force control and visual assessment.
\end{itemize}

These tasks have kept Foxconn's workforce at over 1 million employees and anchored the industry's center of gravity in Shenzhen, where the combination of massive labor pools, component supplier ecosystems, and logistics infrastructure creates powerful agglomeration effects~\cite{krugman1991geography}.

\subsubsection{Capability Analysis}

\textbf{Current state:} $\mathbf{c}_{\text{elec}} \approx (0.70, 0.30, 0.99, 0.40)$.

The dexterity index $\delta = 0.70$ reflects that rigid-component automation is mature but deformable-object manipulation---the core bottleneck---remains in early stages. The generalization index $\gamma = 0.30$ captures the reality that each new product model (e.g., iPhone 16 vs.\ iPhone 17) currently requires 4--8 weeks of line reconfiguration, programming, and validation---a process that keeps human flexibility essential for managing product transitions. Reliability $\rho = 0.99$ is high for existing automated cells but drops significantly when dexterous tasks are attempted. Tactile-vision fusion $\tau = 0.40$ reflects emerging but immature integration of force sensing with visual servoing for connector insertion tasks.

\textbf{Critical threshold for $\Sigma_W$ (weight inversion):} $\delta \approx 0.92$, $\gamma \approx 0.70$, $\rho \geq 0.995$.

The weight inversion occurs when the labor cost weight $w_L$ drops below the combined weight of market proximity $w_M$ and component logistics $w_T$. At $\delta = 0.92$, the remaining human tasks (approximately 6 out of 75 steps) can be handled by a small technical crew rather than a production-scale workforce---eliminating the need for mass labor infrastructure. At $\gamma = 0.70$, product-model transitions that currently take weeks compress to hours, removing the ``retooling penalty'' that favors long production runs at centralized sites.

\subsubsection{Post-Transition Topology}

Beyond $\Sigma_W$, the optimal site-selection equation flips. The dominant cost factor becomes proximity to (a)~consumer markets (reducing logistics cost and lead time for a product with 12--18 month lifecycle) and (b)~semiconductor fabs (reducing component shipping costs and enabling tighter supply chain integration). This favors a distributed topology: regional assembly hubs in North America, Europe, and East Asia, each serving local markets---a radical departure from the current Shenzhen-centric model.

Apple's existing shift of some iPhone assembly to India (targeting domestic sales) can be understood as a proto-signal of this transition, currently driven by geopolitical considerations but structurally enabled by increasing $\delta$ and $\gamma$~\cite{flexport2023electronics}. Once the capability thresholds are fully crossed, the economic case for distributed assembly becomes self-sustaining without geopolitical subsidy.

\textbf{Estimated timeline:} 5--8 years, based on current scaling trends in VLA models~\cite{brohan2023rt2,black2024pi0} and rapid progress in deformable-object manipulation~\cite{zhang2025dexterous,chen2023visual}.

\subsection{Case 2: Aerospace Manufacturing --- The Batch Collapse Trigger}

\subsubsection{Current Landscape}

Aerospace manufacturing occupies a unique position: extremely high value per unit, very low production volumes, and extraordinary precision requirements. Boeing produces approximately 35--45 737 aircraft per month; Airbus delivers around 50--60 A320-family aircraft monthly~\cite{boeing2024deliveries,airbus2024deliveries}. Each aircraft contains approximately 3--6 million individual parts, with final assembly involving tens of thousands of drilling, fastening, sealing, and inspection operations~\cite{campbell2006manufacturing}.

The current manufacturing topology is hyper-concentrated. Boeing's 737 final assembly occurs at a single facility in Renton, Washington; the A320 family is assembled in Toulouse, Hamburg, Tianjin, and Mobile, Alabama. The geographic logic is driven not primarily by labor cost (aerospace workers are highly skilled and well-compensated) but by the enormous \textit{fixed costs} of tooling, jigs, and process validation:

\begin{itemize}[leftmargin=40pt,itemsep=2pt]
    \item A single-aisle aircraft assembly line requires \$2--5 billion in facility and tooling investment~\cite{campbell2006manufacturing}.
    \item Process validation and certification (FAA/EASA) for a new production site takes 2--5 years and costs hundreds of millions of dollars.
    \item Switching between aircraft variants (e.g., 737-800 to 737 MAX 10) on the same line requires weeks of reconfiguration.
\end{itemize}

These fixed costs enforce the current batch-production model: maximize throughput of a single variant before incurring the switching penalty. This is Fordism in its purest surviving form---the moving production line at Renton is a direct descendant of Highland Park~\cite{hounshell1984american}.

\subsubsection{Capability Analysis}

\textbf{Current state:} $\mathbf{c}_{\text{aero}} \approx (0.45, 0.15, 0.997, 0.35)$.

Aerospace automation is surprisingly low relative to the industry's technological sophistication. Dexterity $\delta = 0.45$ reflects that most assembly tasks---particularly those involving confined spaces, overhead work, sealant application, and flexible harness routing---remain manual. The generalization index $\gamma = 0.15$ is extremely low: each automated cell is purpose-built for a specific operation on a specific aircraft model. Reliability $\rho = 0.997$ is high where automation exists (e.g., automated drilling systems) due to the industry's zero-tolerance safety culture. Tactile-vision fusion $\tau = 0.35$ captures emerging uses of force-controlled drilling and automated NDI (non-destructive inspection) but limited integration.

\textbf{Critical threshold for $\Sigma_N$ (batch collapse):} $\gamma \approx 0.75$, $\delta \approx 0.80$, $\rho \geq 0.999$.

The batch collapse pathway is the dominant transition mechanism for aerospace because the industry's central economic constraint is not labor cost but \textit{switching cost and tooling amortization}. The minimum economic batch size (Equation~\ref{eq:mebs}) in aerospace is currently measured in hundreds of aircraft---precisely because $C_{\text{switch}}$ and $C_{\text{facility}}$ are so large.

If foundation models for manipulation achieve $\gamma = 0.75$---enabling a robotic system to adapt from drilling fuselage frames to routing wire harnesses with minimal reconfiguration---then $C_{\text{switch}}$ collapses. If $\delta = 0.80$ is reached, the majority of currently manual tasks become automatable, reducing $C_{\text{labor}}$ and the workforce-infrastructure component of facility cost.

The combined effect: the MEBS drops from hundreds of aircraft to tens. At this point, it becomes economically rational to operate smaller, more distributed final assembly facilities closer to airline customers and MRO (maintenance, repair, overhaul) hubs---a topology that also reduces the catastrophic supply chain risk currently concentrated in single-site production.

\subsubsection{Post-Transition Topology}

Beyond $\Sigma_N$, the aerospace manufacturing map diversifies from a handful of mega-assembly sites to a network of regional assembly hubs. Each hub can flexibly produce multiple aircraft variants, serve regional carriers with reduced delivery logistics, and integrate more tightly with local MRO operations. The current model of shipping wings from Broughton to Toulouse, fuselage sections from Wichita to Renton, and engines from Cincinnati to Everett---a supply chain spanning tens of thousands of miles for a single aircraft---compresses into regional clusters.

This transition also has strategic defense implications. The concentration of military aircraft production at a small number of facilities (e.g., Lockheed Martin's F-35 line in Fort Worth) creates a single point of failure that distributed, highly-automated manufacturing would mitigate~\cite{augustine2023defense}.

\textbf{Estimated timeline:} 10--15 years, constrained by the aerospace industry's conservative certification culture and the extreme reliability requirements ($\rho \geq 0.999$) for safety-critical operations.

\subsection{Case 3: Fresh Food Processing --- The Decoupling Trigger}

\subsubsection{Current Landscape}

Fresh food processing---encompassing fruit and vegetable sorting, meat processing, ready-meal preparation, and bakery production---is a \$4.1 trillion global industry that remains one of the most labor-intensive sectors in high-income economies~\cite{usda2024food}. Unlike electronics or aerospace, where manufacturing is concentrated in developing economies by choice, food processing is often located in high-wage countries by necessity: perishability constraints demand proximity to both agricultural sources and consumers, creating a geographic squeeze that cannot be resolved by offshoring.

The labor challenge is acute. In the United States alone, food processing employs approximately 1.7 million workers~\cite{bls2024food}, with chronic labor shortages driving annual turnover rates exceeding 40\% in meat processing and 30\% in produce handling. The European fresh food sector faces similar pressures, with seasonal labor dependencies creating supply vulnerabilities exposed during COVID-19~\cite{eurofound2021food}.

The manipulation challenges in food processing are uniquely demanding:

\begin{itemize}[leftmargin=40pt,itemsep=2pt]
    \item \textbf{Extreme variability:} Unlike manufactured components, natural products vary continuously in size, shape, weight, ripeness, and surface properties. A bin of strawberries contains no two identical objects.
    \item \textbf{Damage sensitivity:} Bruising thresholds for fresh produce are often below 1~N of localized force~\cite{vanZeebroeck2007bruise}, requiring grasp forces calibrated to individual item properties.
    \item \textbf{Multi-modal quality assessment:} Ripeness, freshness, and defect detection require simultaneous visual (color, surface texture), tactile (firmness, compliance), and sometimes olfactory sensing---a profoundly multi-modal task.
    \item \textbf{Hygiene constraints:} Food-contact surfaces must be washable, chemical-resistant, and compatible with cold/wet environments, severely constraining actuator and sensor design.
\end{itemize}

\subsubsection{Capability Analysis}

\textbf{Current state:} $\mathbf{c}_{\text{food}} \approx (0.40, 0.25, 0.97, 0.20)$.

The dexterity index $\delta = 0.40$ is low because gentle manipulation of highly variable, damage-sensitive objects remains frontier research. Generalization $\gamma = 0.25$ reflects the inability of current systems to transfer skills between product categories (e.g., from apple sorting to fish filleting) without extensive retraining. Reliability $\rho = 0.97$ is moderate---acceptable for many automated steps but insufficient for unsupervised operation in hygiene-critical environments where a single failure (e.g., a dropped product contaminating a line) requires full shutdown and sanitization. Tactile-vision fusion $\tau = 0.20$ captures early-stage work on tactile ripeness sensing and compliant grasping.

\textbf{Critical threshold for $\Sigma_H$ (decoupling):} $\delta \cdot \rho > 0.90$, $\tau \geq 0.70$, $\gamma \geq 0.60$.

The decoupling pathway is uniquely relevant to food processing because of the industry's geographic paradox: it \textit{must} be located near both farms and consumers, but these locations often lack the large, stable labor pools that food processing requires. Rural agricultural regions face depopulation; urban areas have high labor costs and turnover. The industry is caught in a spatial trap that only human-infrastructure decoupling can resolve.

When $\tau$ reaches 0.70---enabling reliable tactile-visual quality assessment across produce categories---and $\delta \cdot \rho$ exceeds 0.90, autonomous food processing becomes viable in locations that currently cannot support the required workforce:

\begin{itemize}[leftmargin=40pt,itemsep=2pt]
    \item \textbf{On-farm processing:} Autonomous sorting, grading, and packaging at the point of harvest, eliminating cold-chain transportation of raw produce to distant processing centers. This alone could reduce food waste by an estimated 15--25\%, as losses during post-harvest transportation are a major contributor to the 30--40\% of food that is wasted globally~\cite{fao2019food}.
    \item \textbf{Urban micro-processing:} Small, autonomous facilities in city centers producing fresh-cut, ready-meal, and bakery products within hours of consumption. The ``dark kitchen'' model, currently dependent on human labor, evolves into fully autonomous food manufacturing co-located with urban demand.
    \item \textbf{Extreme-environment agriculture:} As vertical farming and controlled-environment agriculture expand into arid, Arctic, and space environments~\cite{kozai2015plant}, autonomous food processing becomes the natural complement---closing the farm-to-table loop without requiring human habitation at the production site.
\end{itemize}

\subsubsection{Post-Transition Topology}

Beyond $\Sigma_H$, the food processing map transforms from a network of large centralized plants (each employing 500--2,000 workers) to a dense mesh of autonomous micro-facilities embedded directly into agricultural and urban landscapes. The concept of a ``food processing labor shortage'' becomes meaningless---not because the labor has been replaced in existing facilities, but because the facilities themselves have been redesigned to operate without labor infrastructure.

This has cascading effects on food supply chain architecture. Current cold-chain logistics---designed to ship raw materials from farms to centralized processors to distribution centers to retailers---compresses into hyper-local loops. The \$200+ billion global cold chain logistics market~\cite{grandview2024coldchain} faces structural disruption as the distance between processing and consumption shrinks from hundreds of miles to single digits.

\textbf{Estimated timeline:} 8--12 years for high-value produce (berries, leafy greens); 12--18 years for complex processing (meat, prepared meals), constrained by hygiene certification requirements and the difficulty of achieving $\tau \geq 0.70$ for highly variable natural products.

\subsection{Cross-Case Synthesis}

Table~\ref{tab:cases} synthesizes the three case studies, highlighting the industry-specific pathways and thresholds.

\begin{table}[hbt!]
\centering
\caption{Cross-case comparison of industry-specific phase transition characteristics}
\label{tab:cases}
\small
\begin{tabular}{lccc}
\toprule
& \textbf{Electronics} & \textbf{Aerospace} & \textbf{Food Processing} \\
\midrule
Market size & \$1.1T & \$0.4T & \$4.1T \\
\hline
Current $\mathbf{c}$ & $(0.70, 0.30, 0.99, 0.40)$ & $(0.45, 0.15, 0.997, 0.35)$ & $(0.40, 0.25, 0.97, 0.20)$ \\
\hline
Primary pathway & $\Sigma_W$ (weight inv.) & $\Sigma_N$ (batch collapse) & $\Sigma_H$ (decoupling) \\
\hline
Binding dimension & $\delta, \gamma$ & $\gamma, \delta$ & $\tau, \delta \cdot \rho$ \\
\hline
Key manipulation challenge & Deformable cables & Confined-space assembly & Variable natural objects \\
\hline
Current topology driver & Labor cost & Tooling amortization & Labor availability \\
\hline
Post-transition topology & Regional hubs & Distributed assembly & Hyper-local mesh \\
\hline
Estimated timeline & 5--8 years & 10--15 years & 8--18 years \\
\bottomrule
\end{tabular}
\end{table}

The three cases reveal a common pattern: in each industry, a specific \textit{capability bottleneck} anchors the current manufacturing topology, and a specific \textit{transmission pathway} propagates the removal of that bottleneck into geographic restructuring. The bottleneck is always multi-dimensional (no single capability suffices), and the restructuring is always discontinuous (there is no ``halfway'' between centralized and distributed topology). This supports the phase-transition interpretation: the manufacturing topology is a macroscopic order that emerges from microscopic capability constraints, and it undergoes qualitative reorganization when those constraints cross critical thresholds.

\section{Discussion}

\subsection{Relationship to Existing Literature}

The reshoring literature has extensively documented the role of automation in encouraging domestic production~\cite{deloitte2024reshoring,fratocchi2014reshoring}. However, this literature treats automation capability as an exogenous binary---``automated'' or ``not automated''---without examining the multi-dimensional nature of capability or the nonlinear, threshold-driven mechanism by which capability improvements translate into geographic restructuring. Our framework provides the missing link: a formal mapping from the continuous space of embodied AI capabilities to the discrete topology of manufacturing geography.

The economic geography literature, pioneered by Krugman~\cite{krugman1991geography} and Fujita et al.~\cite{fujita1999spatial}, provides the theoretical foundations for understanding agglomeration and dispersion forces. We extend this framework by introducing embodied AI capability as an endogenous variable that modulates the relative strength of these forces. In particular, we show that capability improvements weaken the agglomeration forces (labor pooling, specialized suppliers) that underpin geographic concentration, while strengthening the dispersion forces (transport cost minimization, market proximity) that favor distributed production.

The embodied AI literature~\cite{duan2022survey,xia2025embodied} has primarily focused on technical capability development without systematic analysis of economic geographic implications. Conversely, the automation economics literature~\cite{acemoglu2019automation,acemoglu2020robots} has focused on labor displacement without considering the spatial reorganization of production. Our work bridges these two streams.

\subsection{Implications for Industrial Policy}

The phase-transition framework has direct implications for national industrial strategy. Countries currently benefiting from labor-cost-driven manufacturing concentration face the prospect of structural decline not through gradual erosion but through sudden phase transition. The policy implication is that adaptation strategies must be developed \textit{before} the critical surface is crossed, as post-transition adjustment may be too disruptive for orderly management.

Conversely, countries with strong energy infrastructure, advanced computational capacity, and proximity to consumer markets may find themselves newly competitive in manufacturing---\textit{not} because they have lowered their labor costs, but because labor cost has become irrelevant. This suggests that the optimal industrial policy for high-wage economies is not to subsidize manufacturing labor but to invest in the capability infrastructure (energy, computation, robotics R\&D) that accelerates the crossing of critical surfaces.

\subsection{Implications for Supply Chain Architecture}

The distributed-manufacturing regime that emerges beyond $\Sigma_N$ implies a fundamental reorganization of global supply chains---from long, linear chains optimized for cost arbitrage to short, networked meshes optimized for responsiveness and resilience~\cite{christopher2011supply}. Inventory management shifts from forecast-driven stockpiling to demand-sensing just-in-time production. Logistics shifts from bulk intercontinental shipping to last-mile delivery. The economic value migrates from physical transportation to digital coordination and capability deployment.

\subsection{Machine Climate Advantage: Case Analysis}

The Machine Climate Advantage (MCA) framework introduced in Section~4.3.1 generates specific, testable predictions about which geographic locations will emerge as next-generation manufacturing hubs in the Phase~III regime. We analyze several candidate regions that score high on $\phi(\mathbf{x})$ despite being historically marginal or absent from the global manufacturing map.

\subsubsection{The Colorado Front Range: A Case Study}

The Denver--Colorado Springs corridor exemplifies Machine Climate Advantage. The region's climate profile reads as a liability for human-centric manufacturing (high altitude at 1,600--1,800m, extreme UV exposure, arid conditions, cold winters) but as a near-ideal specification sheet for autonomous robotic operations~\cite{noaa2024climate}:

\begin{itemize}[leftmargin=40pt,itemsep=2pt]
    \item \textbf{Annual sunshine:} 300+ days per year, among the highest in the continental United States, providing consistent solar energy generation and stable optical conditions for vision systems.
    \item \textbf{Average relative humidity:} 30--45\%, well below the corrosion-acceleration threshold of 60\%~\cite{lee2020corrosion}. Mean time between failures for precision actuators and electronic systems is estimated to increase 1.5--2$\times$ relative to high-humidity environments such as the Gulf Coast or Southeast Asia.
    \item \textbf{Low precipitation:} Average annual rainfall of approximately 400mm (compared to 1,200mm in Shanghai, 2,500mm in Ho Chi Minh City), minimizing weather-related operational disruptions.
    \item \textbf{Thermal stability:} While absolute temperature range is wide ($-15^\circ$C to $+38^\circ$C), the low humidity means that dew point temperature is consistently low, virtually eliminating condensation risk on cold-start equipment---a significant reliability factor for optical sensors and tactile arrays.
    \item \textbf{Renewable energy potential:} The Front Range corridor has among the highest combined solar and wind energy potential in the United States~\cite{nrel2024atlas}, supporting the energy-autonomous factory model.
    \item \textbf{Logistics:} Central continental location equidistant from both coasts, with established highway and rail infrastructure, provides a natural logistics hub for distributed domestic manufacturing.
\end{itemize}

Denver has historically punched well below its economic weight as a manufacturing hub. Despite being the largest metropolitan economy in the Mountain West, its manufacturing sector is dwarfed by those of comparably sized metros with port access or heavy-industry legacy (e.g., Houston, Detroit, or the Research Triangle). The region's existing manufacturing---primarily aerospace components (Lockheed Martin Space, Ball Aerospace), food processing, and medical devices---is niche and high-value, but the Front Range has never attracted the large-scale, labor-intensive production that defines traditional manufacturing powerhouses~\cite{bls2024food}. Under Phase~I logic, the reasons are clear: no deep-water port, no legacy of heavy industry, and a labor market oriented toward technology and services rather than production-line work. Under Phase~III logic, every one of these ``disadvantages'' becomes irrelevant---and the region's high $\phi(\mathbf{x})$ score makes it a candidate for autonomous manufacturing clusters that conventional site-selection models would never have identified.

\subsubsection{Global MCA Hotspots}

Extending this analysis globally, several regions emerge as high-MCA locations that are currently underrepresented in manufacturing:

\textbf{The Atacama--Altiplano region (Chile/Argentina):} The driest non-polar desert on Earth, with relative humidity often below 15\%, extraordinary solar irradiance ($>$2,500 kWh/m$^2$/year), and vast lithium and copper deposits. Already host to the world's most advanced astronomical observatories (which share the machine-optimal requirement of low humidity, low dust, and clear skies), this region could become a node for autonomous mineral processing and advanced manufacturing.

\textbf{Interior Australia:} Arid, low-humidity, high-irradiance conditions across vast areas, combined with proximity to mineral resources (iron ore, rare earths, bauxite). Currently, the remoteness and harsh climate make human-staffed manufacturing uneconomical; autonomous factories would face no such constraint.

\textbf{The Arabian Peninsula interior:} Regions away from the humid coastal strip offer extreme aridity, near-constant sunshine, and access to the abundant capital and energy infrastructure of Gulf states already investing heavily in post-oil economic diversification. Saudi Arabia's NEOM project~\cite{neom2024report}, while currently conceived around human habitation, could pivot toward autonomous manufacturing zones that exploit the region's MCA.

\textbf{Iceland and Greenland:} Abundant geothermal and wind energy, cool temperatures (favorable for electronics thermal management), low humidity in interior regions, and strategic North Atlantic location between North American and European markets. Currently unviable for manufacturing due to tiny populations; MCA-driven siting eliminates this constraint.

\subsubsection{The Orthogonality Principle}

A striking pattern emerges from this analysis: the locations with the highest Machine Climate Advantage are \textit{orthogonal} to current manufacturing density. This is not because high-MCA locations are necessarily inhospitable to humans---Denver is one of America's most desirable cities to live in---but because the environmental factors that benefit machines (aridity, solar irradiance, thermal stability) have been \textit{completely absent} from the traditional manufacturing site-selection calculus. For over a century, factories were placed by optimizing over labor cost, labor availability, port access, and proximity to supply chains~\cite{weber1929theory,krugman1991geography}. Humidity, solar irradiance, and dust levels simply did not appear in the objective function, because they did not affect the performance of the human workers who staffed the factories.

This orthogonality means that high-MCA locations span a wide spectrum of human livability. Some---such as the Colorado Front Range, the U.S. Desert Southwest, or inland Iberia---are livable and even thriving urban regions, yet their manufacturing presence has historically been modest relative to their economic size, because the traditional siting calculus prioritized factors they lack: deep-water ports, dense blue-collar labor pools, or proximity to legacy industrial clusters. Others---the Atacama Plateau, central Australia, or high-latitude Iceland---have sparse human populations and enter the feasible manufacturing set only through full human-infrastructure decoupling. What unites these diverse locations is not a shared hostility to human life, but a shared possession of environmental attributes---aridity, solar abundance, atmospheric stability---that \textit{one hundred years of manufacturing site-selection theory has simply never weighted}.

The Phase~III transition therefore does not merely redistribute manufacturing within the existing set of industrialized regions. It opens an entirely new geography of production that has \textit{no precedent} in the history of manufacturing---not in the Fordist era, not in the offshoring era, and not in the reshoring era. It is the emergence of a manufacturing map drawn by the logic of machines rather than the logic of labor markets.

\subsection{Limitations and Future Work}

This paper establishes a conceptual and mathematical framework; empirical validation remains future work. Several important limitations apply: (1) the critical threshold values proposed are order-of-magnitude estimates that require systematic experimental validation; (2) the model assumes that capability improvements in $\mathcal{C}$ are monotonic and irreversible, which may not hold in practice due to regulatory, social, or geopolitical barriers to adoption; (3) the phase-transition analogy, while illuminating, may oversimplify the dynamics of transition---real economic systems exhibit hysteresis, path dependence, and multi-stable equilibria not fully captured by our framework; (4) we do not model the feedback loop by which geographic restructuring itself affects the incentives for further capability development.

Future work should: (a) develop empirical methods for measuring the capability indices $(\delta, \gamma, \rho, \tau)$ across industries; (b) estimate critical threshold values through combined technical assessment and economic modeling; (c) analyze the dynamics of transition---including adjustment costs, stranded assets, and labor market disruption---for specific industries and regions; (d) explore the interaction between embodied AI capability thresholds and other megatrends (decarbonization, geopolitical fragmentation, demographic aging) that are simultaneously reshaping manufacturing geography.

\section{Conclusion}

We have argued that the transformative potential of embodied intelligence in manufacturing extends far beyond efficiency gains within existing production systems. By defining a formal Capability Space and identifying three transmission pathways through which capability improvements propagate into structural economic change, we show that embodied AI is poised to trigger \textit{phase transitions} in manufacturing economic geography---discontinuous reorganizations of where, how, and at what scale production occurs.

The framework of Embodied Intelligence Economics provides a principled basis for answering questions that neither the robotics community nor the economics community can answer alone: When will garment manufacturing leave Bangladesh? What capability level makes Arctic manufacturing viable? Why might Denver or the Atacama Desert become manufacturing hubs in a world of autonomous factories? How will foundation models for manipulation reshape the $\$10$ trillion global logistics industry?

These are not questions about efficiency. They are questions about the topology of the global economic system. Answering them requires the kind of interdisciplinary analysis---bridging robotic manipulation, foundation models, economic geography, and supply chain theory---that this paper aims to initiate.

The message for researchers, policymakers, and industry strategists is clear: embodied intelligence is not just making factories faster. It is about to change where factories \textit{exist}---for the first time since Henry Ford changed the answer to that question in 1913.

\bibliographystyle{plain}
\bibliography{references}

\end{document}